\documentclass[11pt]{article}

\usepackage[preprint]{acl}
\usepackage{times}
\usepackage{latexsym}
\usepackage[T1]{fontenc}
\usepackage[utf8]{inputenc}
\usepackage{microtype}
\usepackage{inconsolata}
\usepackage{booktabs}
\usepackage{amsmath}
\usepackage{url}
\usepackage{graphicx}

\title{Gaze as Evidence for Common Grounding: A Cross-Corpus Analysis of MapTask and MUNDEX}

\author{Nan Li\textsuperscript{1}, Albert Gatt\textsuperscript{1}, Massimo Poesio\textsuperscript{1,2}\\
        \textsuperscript{1}Utrecht University, Utrecht, The Netherlands\\
        \textsuperscript{2}Queen Mary University of London, London, The United Kingdom\\
        \texttt{\{n.li, a.gatt, m.poesio\}@uu.nl}}

\begin{document}
\maketitle

\begin{abstract}
In collaborative tasks with asymmetric information, participants coordinate their understanding through interaction.
We ask whether gaze provides evidence about grounding across two such tasks.
Working from discrete behavioral annotations, we map HCRC MapTask~\citep{anderson1991hcrc} and MUNDEX~\citep{turk2023mundex} into a shared partner/task/away vocabulary and compute gaze features around task-relevant dialogue units.
In both corpora, aligned reference interpretations (MapTask) and UND (understood) judgments (MUNDEX) are associated with more task-directed gaze and with less partner-directed gaze, lower gaze entropy, and fewer gaze transitions.
The associations are clearest for the participant leading the task: in giver-produced references, and in explainer judgments, which also co-vary with the explainee's gaze.
In same-speaker MapTask reference chains, the speaker's gaze entropy is lower at the mention where a previously non-aligned referent becomes aligned.
The best gaze feature groups improve modestly over controls under grouped cross-validation: temporal features in MapTask and raw proportions in MUNDEX.
Because effects are small and several weaken when recurring participants rather than dialogues are the unit of inference, we treat gaze as one contributing cue to grounding, to be interpreted alongside task and dialogue context.
\end{abstract}

\section{Introduction}

In collaborative tasks where participants hold different private information, mutual understanding cannot be assumed from shared context alone. It must be built and tracked through interaction~\citep{clark1986referring,clark1991grounding}. Gaze is an observable cue to this process: participants look at task materials, at each other, or away while giving instructions, checking understanding, and coordinating their perspectives.
Some corpora annotate gaze from video as discrete categories of where participants look, rather than as eye-tracking coordinates. These annotations can be used to study the relationship between gaze and grounding, but they are often corpus-specific, making it difficult to compare across tasks.

We study two settings where information asymmetry forces participants to continuously coordinate understanding.
In HCRC MapTask~\citep{anderson1991hcrc}, a giver and a follower navigate with maps that differ in their landmarks; perspectivist grounding labels record each participant's interpretation separately~\citep{li2026grounded}. In MUNDEX~\citep{turk2023mundex}, an explainer teaches a board game to an explainee; both annotate the explainee's moment-by-moment understanding retrospectively.

Building on within-corpus studies, we compare grounding-related gaze patterns across tasks. We contribute (1)~a shared partner/task/away representation that maps two gaze ontologies and applies to other corpora with video-coded gaze annotations; (2)~evidence of directional convergence across distinct grounding measures, clearest for the participant leading the task; and (3)~a within-speaker reference-chain analysis showing that speaker gaze entropy is lower when a referent becomes aligned. Grouped prediction and role-stratified tests characterize the strength and scope of these associations.

\begin{figure*}[t]
\centering
\includegraphics[width=0.92\textwidth]{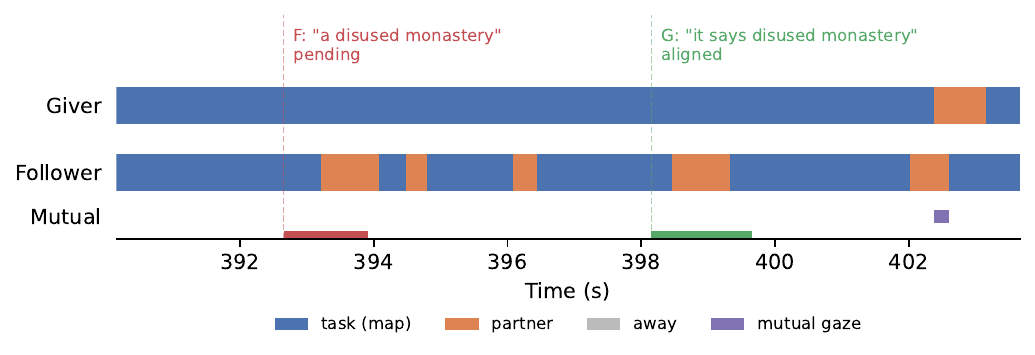}
\caption{MapTask dialogue q8ec4 with gaze mapped to the shared partner/task/away vocabulary. Each horizontal bar is a gaze event; dashed vertical lines mark two reference expressions for the same landmark. The follower's reference to \emph{a disused monastery} is \emph{pending} (red dashed line): the follower repeatedly glances at the partner (orange bars). After the giver's re-mention, the reference is \emph{aligned} (green dashed line). Both participants' gaze remains predominantly task-directed (blue bars).}
\label{fig:timeline}
\end{figure*}

\section{Related Work}

Grounding theory holds that interlocutors seek and provide evidence of understanding as a conversation progresses~\citep{clark1986referring,clark1991grounding}. Visual evidence is part of this process: when directors could see builders' workspace in a Lego assembly task, builders displayed understanding through actions, gaze, and head gestures, and directors adjusted their utterances accordingly~\citep{clark2004monitoring}. Gaze also carries referential information, as matchers used a director's gaze to identify targets before the linguistic point of disambiguation~\citep{hanna2007gaze}.

Map-based tasks tie gaze more directly to grounding complexity. In MapTask dialogues with visibility, followers looked up at givers more often while discussing landmarks that differed between their maps~\citep{boyle1994effects}. In a direction-giving study, \citet{nakano2003face} coded gaze at the partner, the map, and elsewhere, and found that nonverbal patterns differed by dialogue act: after a giver's assertion, a listener's sustained gaze at the speaker was usually followed by elaboration, whereas continued attention to the map more often preceded the next instruction. \citet{murat2026gaze} aligned both participants' MapTask gaze with turn boundaries and related it to dialogue acts, lexical entropy, and repetition; partner-directed gaze at turn ends accompanied turns expressing difficulty, while map-directed gaze was more typical of exchanges without obstacles or disagreement.

In MUNDEX, understanding was annotated through retrospective video recall~\citep{turk2023mundex}. \citet{wang2026predicting} related explainees' self-reported understanding to speaker information value, syntactic complexity, and listener gaze entropy, computed as the average negative log-probability of automatically estimated gaze labels under a sequence model; adding these cues to textual features improved classification. \citet{lazarov2026gaze} manually coded MUNDEX gaze as directed to the partner, the table, or away; in the explanation phase without the board game, gaze aversions to the table or away were associated with topic changes.

The perspectivist MapTask annotation records speaker and addressee interpretations separately~\citep{li2026grounded} and has been used to evaluate whether vision-language models track common ground~\citep{li2026seeing}. Its reference-level labels allow gaze to be compared across repeated mentions of the same landmark. We map both corpora into one partner/task/away vocabulary and compare gaze associations across tasks and grounding measures.

\section{Data and Representation}
\label{sec:data}

\paragraph{MapTask}
We use the gaze-annotated portion of HCRC MapTask~\citep{anderson1991hcrc} with perspectivist labels from \citet{li2026grounded}, where a reference expression is \emph{aligned} only when speaker and addressee interpretations resolve to the same landmark. Dialogues come in an eye-contact condition (ec), where participants can see each other, and a no-eye-contact condition (nc); the \emph{up}$\rightarrow$partner mapping is only literally partner-directed in the ec condition.

Of the corpus's 94 gaze files, 46 dialogues (31 ec, 15 nc) have gaze annotations for both participants; we match grounding annotations to landmark-reference annotations by dialogue, role, and landmark identity. After filtering windows where either participant has less than 30\% gaze coverage (17 windows ruled out), we obtain 5,144 reference-expression windows: 3,807 aligned, 1,261 pending, and 76 misunderstood.

\paragraph{MUNDEX}
MUNDEX~\citep{turk2023mundex} records explainers (EX) teaching a board game to explainees (EE) in German. After each task, participants watched their recording: EE reported their own understanding and EX judged EE's understanding on a four-level scale: \emph{understood} (UND), \emph{partially understood} (PART\_UND), \emph{not understood} (NON\_UND), and \emph{misunderstood} (MISUND).
We combine EX judgments and EE self-reports in a pooled analysis of \emph{annotator-judged understanding}, retaining each annotation as a separate observation with its own gaze window. In the 26 interactions with both gaze tiers, 956 valid annotations (524 EX, 432 EE) yield 807 windows (458 EX, 349 EE) after the 30\% coverage filter (149 excluded): 360 UND, 199 PART\_UND, 151 NON\_UND, and 97 MISUND.

\paragraph{Gaze representation}
Both corpora annotate gaze as discrete behavioral categories from video, not as eye-tracking coordinates. We map both into a shared partner/task/away vocabulary: MapTask's \emph{up}/\emph{down}/\emph{off} become partner/task/away; MUNDEX's EX/EE/TABLE/AWAY map analogously. Figure~\ref{fig:timeline} shows a MapTask excerpt with both participants' mapped gaze and two reference expressions for the same landmark. We discard gaze events with non-positive duration (annotation noise) and resolve temporal overlaps within each participant's gaze stream. Full corpus details and window definitions are in Appendix~\ref{app:corpora}; the features computed from this representation are described in Section~\ref{sec:experiments}.

\section{Experiments}
\label{sec:experiments}

\paragraph{Features}
From the shared partner/task/away vocabulary we compute gaze features per window in several groups (full definitions in Appendix~\ref{app:features}). \emph{Raw proportions} record how much of the window each participant spends on each gaze target, plus mutual gaze (7 features in MapTask, 8 in MUNDEX). The \emph{structured} set adds coverage, transition count, and Shannon entropy
\footnote{Here, we use a duration-weighted Shannon entropy: $H=-\sum_k q_k\log_2 q_k$, where $q_k$ is the proportion of observed gaze time directed toward canonical target $k\in\{\text{partner, task, away}\}$ within the analysis window.} (13/14 features total). Four further groups capture finer-grained patterns: (1)~\emph{temporal dynamics} capturing the timing of gaze shifts: gaze-run counts, durations, switch rate, latency, and first/last/dominant-label indicators (21 per participant); (2)~\emph{transition bigrams} encoding the direction of gaze switches: proportions of ordered label pairs such as task$\rightarrow$partner (6 per participant); (3)~\emph{coordination} measuring whether both participants' gaze is synchronized: joint gaze states sampled at approximately 10\,Hz (at least 20 points per window), namely mutual task and partner gaze, gaze alignment, complementary gaze, joint entropy, and partner coupling (6 joint features); and (4)~\emph{derived ratios} expressing relative gaze preferences: partner/task ratio, engagement, task dominance, and between-participant asymmetries (9 features).

\paragraph{Association and process analyses}
We use the features defined above to test whether gaze patterns differ between grounding states. Binary contrasts (Mann--Whitney U with rank-biserial correlations) compare aligned vs.\ non-aligned windows in MapTask and annotator-judged UND vs.\ non-UND in MUNDEX, stratified by interactional role and, in MapTask, by eye-contact condition, and assessed with cluster-robust logistic GEE~\citep{liang1986longitudinal}; $q$ values are Benjamini--Hochberg (BH) adjusted~\citep{benjamini1995controlling}. We also track gaze across repeated mentions of the same landmark within each dialogue to assess gaze change around alignment within speakers. Section~\ref{sec:results} reports the main results (Table~\ref{tab:associations}; Figure~\ref{fig:resolution}), and Appendices~\ref{app:associations}--\ref{app:chains} give the full tests.

\paragraph{Prediction setup}
We also test whether the gaze features carry recoverable signal through a simple prediction task. We binarize the corpus-specific labels (aligned vs.\ non-aligned in MapTask; annotator-judged UND vs.\ non-UND in MUNDEX) because minority classes are small after intersecting with gaze coverage. All models are logistic regression (LR) with standardized features and balanced class weights.
We evaluate with grouped cross-validation: 10-fold grouped by dialogue for MapTask and 5-fold grouped by explainer for MUNDEX, evaluating on held-out dialogues or explainers. We ablate each feature group and their combinations. Section~\ref{sec:results} reports the results (Table~\ref{tab:model-results}), and Appendix~\ref{app:prediction} gives implementation details and full results.

\section{Results}
\label{sec:results}

\begin{figure*}[!t]
\centering
\includegraphics[width=0.82\textwidth]{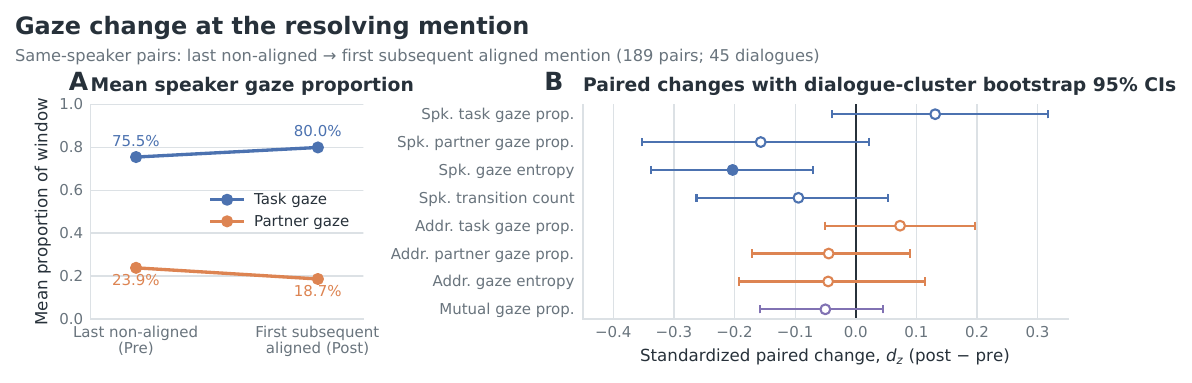}
\caption{Within-chain gaze change at the first subsequent aligned mention (189 same-speaker pairs from 45 dialogues). Panel~A: mean speaker task- and partner-gaze proportions before (Pre) and after (Post) resolution; lines connect sample means, not individual trajectories; away gaze is not shown (mean ${<}$1.5\%). Panel~B: standardized paired changes ($d_z$\,=\,mean(Post$-$Pre)\,/\,SD; paired Wilcoxon, BH-corrected) for all eight tested features; error bars are dialogue-cluster bootstrap 95\% CIs (resampling dialogues to respect within-dialogue dependence); filled marker indicates $q{<}.05$. Only speaker gaze entropy survives correction ($q{=}.044$); dialogue-level aggregation does not ($q{=}.20$).}
\label{fig:resolution}
\end{figure*}

\begin{table}[t]
\centering
\footnotesize
\setlength{\tabcolsep}{3pt}
\begin{tabular}{llrr}
\toprule
Corpus & Feature & $\Delta$ & $r$ \\
\midrule
MapTask & speaker task gaze & .033 & .058 \\
MapTask & speaker partner gaze & -.029 & -.057 \\
MapTask & speaker entropy$^\dagger$ & -.046 & -.054 \\
MapTask & speaker transitions$^\dagger$ & -.091 & -.053 \\
MUNDEX & EX task gaze$^\dagger$ & .105 & .181 \\
MUNDEX & EX partner gaze$^\dagger$ & -.102 & -.172 \\
MUNDEX & EE partner gaze$^\dagger$ & -.084 & -.148 \\
MUNDEX & mutual gaze explicit$^\dagger$ & -.060 & -.143 \\
\bottomrule
\end{tabular}
\caption{Four largest associations per corpus, selected by $|r|$. $\Delta$: positive-class mean minus negative-class mean; $r$: rank-biserial correlation. All Mann--Whitney $p{<}.001$; $^\dagger$ also significant under cluster-robust GEE (FDR $q{<}.05$; Appendix~\ref{app:gee}). Full results in Appendix~\ref{app:associations}.}
\label{tab:associations}
\end{table}

\paragraph{Associations and roles}
Table~\ref{tab:associations} shows the four strongest associations per corpus (full results in Tables~\ref{tab:app-maptask-mw}--\ref{tab:app-mundex-mw}). In both corpora, the largest associations point the same way: task-gaze proportions are higher and partner-gaze proportions lower for the positive class, while entropy and transitions tend to be higher for the negative class. Effects are small: the largest pooled $|r|$ is .058 in MapTask and .181 in MUNDEX.

In MapTask, speaker task- and partner-gaze proportions are significant in these window-level tests but not under dialogue-clustered GEE ($q{=}.091$); speaker entropy and transitions remain significant. With recurring participants as clusters and bias-reduced standard errors, four MUNDEX associations remain significant: the explainer's task- and partner-gaze proportions ($q{=}.001$ and $q{=}.028$) and the explainee's entropy and transitions (both $q{=}.039$); no MapTask feature does (Appendix~\ref{app:gee}).

In MapTask, associations are clearest for giver-produced references (six significant features; largest $|r|$ .086), whereas follower-produced references show near-zero effects (largest $|r|$ .043; Table~\ref{tab:app-maptask-roles}).
In MUNDEX, all 14 structured features have rank-biserial correlations of the same sign in EX judgments (458 windows) and EE self-reports (349). The largest effect is greater for EX judgments ($|r|$ .206 vs.\ .151), the only stratum with features surviving correction; these include the explainee's gaze proportions, entropy, and transitions (Table~\ref{tab:app-mundex-roles}).
UND is the task-directed extreme across all four gaze measures, though the remaining understanding classes do not follow a consistent order (Table~\ref{tab:app-mundex-kruskal}).

All 13 MapTask features have larger $|r|$ in the eye-contact stratum than in the pooled data (largest .078 vs.\ .058), whereas none is significant in the no-eye-contact stratum (largest $|r|$ .025), where partner-directed gaze is largely absent; the formal condition interaction is not significant (Table~\ref{tab:app-maptask-conditions}).

\paragraph{Gaze across the grounding process}
Reference chains group repeated mentions of the same landmark within a dialogue. Across chain positions, the aligned rate rises from .30 at first mentions to .59 at second mentions and .85 in the fourth-and-later bucket, and mean speaker partner gaze, entropy, and transitions are lower at second than at first mentions (Table~\ref{tab:app-chain-positions}).

To examine change within chains, we compare the speaker's gaze at the last non-aligned mention with gaze at the resolving aligned mention, restricting to within-speaker pairs where the same person produced both ($n{=}189$ pairs in 45 dialogues). Speaker entropy decreases significantly after BH correction ($d_z{=}{-}.20$, $q{=}.044$), and its dialogue-cluster bootstrap 95\% CI excludes zero (Figure~\ref{fig:resolution}; Table~\ref{tab:app-resolution}); partner gaze, task gaze, and transitions shift in the same directions but do not survive correction. The decrease depends on the inference unit: it does not survive correction when pair differences are averaged within each of the 45 dialogues ($q{=}.20$), and resampling the six groups of dialogues that share participants yields a CI below zero, but the decrease is concentrated in two of these groups (Appendix~\ref{app:chains}). Because later mentions are both more often aligned and more task-directed, we also compare aligned and non-aligned mentions at the same chain position; only second-mention task gaze survives correction ($q{=}.01$; Table~\ref{tab:app-position-matched}).

\begin{table}[t]
\centering
\small
\begin{tabular}{lcc}
\toprule
Feature set (macro-F1) & MapTask & MUNDEX \\
\midrule
Majority baseline & .425 & .356 \\
LR controls only & .472 & .544 \\
LR raw proportions & .522 & \textbf{.564} \\
LR structured & .512 & .543 \\
LR structured+temporal & \textbf{.532} & .550 \\
LR structured+coordination & .511 & .551 \\
LR all extended & .527 & .527 \\
\bottomrule
\end{tabular}
\caption{Feature-group ablation (macro-F1). LR = logistic regression. ``Controls only'' uses condition (ec/nc) and speaker role in MapTask, and annotator role (EX/EE) in MUNDEX.}
\label{tab:model-results}
\end{table}

\paragraph{Prediction and ablation}
The two corpora favor different feature groups: structured+temporal features give the highest macro-F1 in MapTask (.532) and raw proportions in MUNDEX (.564; Table~\ref{tab:model-results}). These exceed controls-only scores by .060 and .020, respectively; MUNDEX structured gaze (.543) does not exceed its role-only control (.544). Across 30 reshuffled grouped partitions, these groups score highest in 27 partitions in each corpus. The gains remain modest and partition-dependent: the MapTask gain over controls ranges from .015 to .070 (mean .038), and the MUNDEX gain is positive in 29 partitions and at most .027 (Appendix~\ref{app:prediction}). Within the EE self-report stratum, however, four of six engineered groups score above raw proportions (Appendix~\ref{app:mundex-perspectives}).

\section{Discussion}

\paragraph{Shared categories, task-specific meanings}
The direction of these associations is the same in both corpora, echoing map-task observations that partner-directed gaze increases around communicative difficulty~\citep{boyle1994effects,nakano2003face,murat2026gaze}. The two labels measure different constructs: MapTask records referential alignment, whereas MUNDEX pools explainees' self-reports and explainers' judgments, so the convergence spans related but distinct grounding measures.

Which features carry predictive signal differs: temporal dynamics score highest in MapTask and raw proportions in MUNDEX.
The shared categories also name gaze targets rather than functions. In MapTask, a partner glance may check a landmark reference; in MUNDEX, gaze averted from the partner has also been linked to topic changes~\citep{lazarov2026gaze}, so it may organize an explanation as well as reflect understanding. Comparing finer-grained referents and dialogue actions would help distinguish task-general patterns from task-specific behavior.

\paragraph{Gaze and interactional role}
Significant associations concentrate in giver-produced references and explainer judgments, whereas follower-produced references show near-zero effects. This pattern fits the task structure: givers produce the instructions being grounded, and explainers monitor explainees, whose gaze proportions and dynamics co-vary with explainers' judgments. Feature$\times$role interactions do not survive correction (MapTask $q{=}.064$; not significant in MUNDEX; Appendix~\ref{app:roles}), so we report a stratum difference rather than a tested moderation, and role-conditioned modeling remains to be tested.

\paragraph{Implications}
Because the representation uses discrete behavioral categories rather than eye-tracking coordinates, it can be applied to other corpora with video-coded gaze annotations, making cross-corpus comparisons of grounding behavior easier to set up. The associations and the prediction gains over controls indicate grounding-related signal in gaze that is worth modeling together with lexical content, dialogue acts, and task state, and testing across corpora.

\section{Conclusion}

In collaborative tasks with asymmetric information, gaze provides directionally consistent evidence about grounding. After mapping MapTask and MUNDEX annotations into a shared partner/task/away vocabulary, aligned references and UND judgments are both accompanied by a higher share of task gaze, a lower share of partner gaze, and fewer gaze transitions, most clearly in giver-produced references and explainer judgments. In MapTask reference chains, the speaker's gaze entropy is lower at the mention where a referent becomes aligned. These effects are small and partly depend on the unit of inference; gaze should therefore be modeled as part of the interactional state, alongside linguistic and task-context features.

\section{Limitations}

\paragraph{Representation and labels}
Three gaze categories cannot identify the specific landmark or object being viewed. Shared category names do not establish equivalent interactional functions across tasks. MapTask's up$\rightarrow$partner mapping is literal only with eye contact; associations are detected in that stratum, but the condition interaction is not significant. MUNDEX's retrospective self-reports and partner judgments measure different perspectives, and linked events can contribute conflicting labels: 271 of 807 rows belong to reconstructed two-perspective links, and 44 of the 135 fully retained pairs disagree on the binary label. Retaining one row per pair preserves the positive association between the explainer's task-gaze proportion and UND (Appendix~\ref{app:mundex-perspectives}).

\paragraph{Evidence and scope}
The data comprise 46 MapTask dialogues and 26 MUNDEX interactions with nine explainers, and participants recur: the MapTask dialogues involve 24 participants in six groups connected by shared participants, and eight of the nine MUNDEX explainers take part in three interactions. With these groups or explainers as GEE clusters and bias-reduced standard errors, no MapTask feature survives correction; in MUNDEX, explainer task and partner gaze and explainee entropy and transitions remain significant, whereas explainee gaze proportions and mutual gaze do not (Appendix~\ref{app:gee}). The modest effects, small number of groups, sensitivity of the chain result to the inference unit, and sensitivity of prediction gains to the cross-validation partition limit conclusions about dynamic grounding and predictive generalization. Coverage filtering also restricts which moments enter the analysis, and unevenly so: two MUNDEX explainers account for 139 of the 149 excluded windows. Two asymmetric, face-to-face tasks in English and German support directional convergence; other task structures and multimodal predictors remain to be tested.

\section*{Acknowledgments}

We appreciate the helpful comments and suggestions from the anonymous reviewers.
This work is funded by the Dutch Research Council (NWO) through the AiNed Fellowship Grant NGF.1607.22.002, \textit{Dealing with Meaning Variation in NLP}.

\section*{Ethics Statement}

This work uses only publicly released corpora, HCRC MapTask and MUNDEX. We analyze their annotations; the MapTask and MUNDEX annotations do not identify individual participants.

\section*{Data and Code Availability}

All source data are publicly available. The HCRC MapTask gaze, timing, and landmark-reference annotations come from the NXT release 2.1 of the corpus (CC BY 4.0).\footnote{\url{https://groups.inf.ed.ac.uk/maptask/maptasknxt.html}} The perspectivist grounding labels \citep{li2026grounded} are released as the Grounded Misunderstandings in MapTask (GMMT) dataset (CC BY 4.0).\footnote{\url{https://github.com/chnln/grounded-misunderstandings-in-maptask}; \url{https://huggingface.co/datasets/chnln/grounded-misunderstandings-in-maptask}} The MUNDEX annotations \citep{turk2023mundex} are available from Zenodo (version 0.7; CC BY 4.0).\footnote{\url{https://doi.org/10.5281/zenodo.17129817}} MUNDEX audio and video are not released. Our processed gaze feature tables and the analysis code are available in our public repository.\footnote{\url{https://github.com/chnln/gaze-as-grounding-evidence}}

\bibliography{references}

\appendix

\section{Corpora, Annotations, and Preprocessing}
\label{app:corpora}

\subsection{HCRC MapTask}

In HCRC MapTask \citep{anderson1991hcrc}, an instruction giver guides a follower along a route using maps that deliberately differ in their landmarks. Dialogues come in an eye-contact condition (ec), where participants can see each other, and a no-eye-contact condition (nc). Gaze is annotated per participant with three categories from the corpus ontology: \emph{up} (looking up, toward the partner when eye contact is possible), \emph{down} (looking at the map), and \emph{off} (looking off camera). The corpus release contains 94 gaze XML files; 46 dialogues (31 ec, 15 nc) have gaze for both participants and constitute our analysis set.

Gaze events, word timings, and landmark references live in three separate NXT annotation layers that do not reference each other directly. A landmark reference lists word IDs, so we resolve its start and end times through the timed-units layer. This yields 5,162 timed landmark references across the 46 dialogues.

The grounding labels come from the perspectivist re-annotation of \citet{li2026grounded}, which labels each reference expression (RE) as \emph{aligned}, \emph{pending}, or \emph{misunderstood} depending on whether speaker and addressee interpretations resolve to the same landmark. We match grounding annotations to timed landmark references per dialogue and speaker, in order, by landmark ID; all 5,161 grounding annotations in the 46 dialogues match (100\%). After the coverage filter (Appendix~\ref{app:windows}), 5,144 windows remain: 3,807 aligned, 1,261 pending, 76 misunderstood; 3,199 ec and 1,945 nc; 3,446 giver-produced and 1,698 follower-produced.

Each observation represents an RE--landmark pair, so expressions referring to multiple landmarks can contribute observations with identical gaze windows (54 observations across 26 windows). Four windows contain both aligned and non-aligned labels for different landmarks. We test two ways of retaining one observation per window: assigning a non-aligned label if any observation in the window is non-aligned, or excluding the four conflicting windows and retaining one observation per remaining window. Both preserve the sets of significant structured features and change gaze-model macro-F1 by at most .004, although controls-only macro-F1 rises from .472 to .498--.499; the same-speaker chain pairs are unaffected.

\subsection{MUNDEX}

MUNDEX \citep{turk2023mundex} records explainers (EX) teaching a board game to explainees (EE) in German. Afterwards, participants watched their own recording and retrospectively judged the explainee's understanding (video-recall), giving the tiers EE\_UND (explainee's own understanding) and EX\_UND (explainer's judgment of the explainee's understanding) with values UND, PART\_UND, NON\_UND, and MISUND.
Gaze tiers give each participant's gaze target (the interlocutor, TABLE, or AWAY), and a MUTUAL\_GAZE tier marks mutual gaze explicitly. The release contains 45 ELAN files; 26 interactions (9 explainers) have both gaze tiers and form our analysis set. The 26 interactions contain 956 valid judgments (524 EX, 432 EE); filtering excludes 149 windows (66 EX, 83 EE). The remaining 807 annotator-judged understanding windows comprise: 360 UND, 199 PART\_UND, 151 NON\_UND, 97 MISUND; 349 explainee-annotated and 458 explainer-annotated.

\subsection{Understanding perspectives and linked annotations}
\label{app:mundex-perspectives}

The MUNDEX target is annotator-judged understanding: UND is positive and PART\_UND, NON\_UND, and MISUND are negative.
EX judgments of EE and EE self-reports remain separate annotation rows.

We reconstruct links from UND\_MATCH=YES spans by collecting valid EX and EE annotations whose starts lie within each span expanded by 60\,ms at both ends, and retaining a link only when there is exactly one candidate per perspective. Of 164 spans in gaze-complete interactions, 159 give clean links; 135 retain both rows after coverage filtering and one retains only one row, accounting for 271/807 rows (33.6\%). Among the 135 retained pairs, 58 agree on the four-class label and 91 on the binary label; 44/135 (32.6\%) conflict. Their median anchor separation is 4.36\,s, so the paired windows need not coincide.

As a sensitivity check, retaining the EX, EE, or a seeded random row from each pair yields 672 rows. Across these variants, 11--12 of 14 structured features survive within-variant BH correction, compared with 11/14 pooled; the explainer's task-gaze proportion remains positively associated with UND ($r=.201$--$.229$, pooled $r=.181$). Interaction-clustered GEE and held-out-explainer cross-validation keep linked observations within the same inference cluster and fold, respectively.

Perspective-specific prediction uses the same five-fold held-out-explainer procedure and the eight-feature raw gaze group. Raw-gaze macro-F1 is .549 versus a within-stratum majority baseline of .337 for EX judgments ($n=458$), and .520 versus .380 for EE self-reports ($n=349$); pooled scores are .564 versus .356 ($n=807$). The EX/EE role indicator alone gives pooled macro-F1 .544; it is constant within a perspective. Feature-group rankings also differ by perspective: raw gaze is the highest-scoring group for EX judgments, whereas four of six engineered groups exceed it for EE self-reports (highest: structured+ratios, .556). These are point estimates from overlapping subsets of the same nine explainers.

\subsection{Windows, cleaning, and coverage}
\label{app:windows}

MapTask windows span a reference expression plus 1.5\,s of post-context, chosen to capture the addressee's immediate reaction (confirmation, repair initiation, or gaze shift) following the expression. MUNDEX windows span an understanding annotation $\pm$2\,s, clipped at zero; understanding annotations are short intervals (typically 1\,s), so symmetric context is needed to capture gaze dynamics around the annotated moment.
Gaze events with non-positive duration are discarded. For structured features, time covered by overlapping events of the same participant is counted once, so observed time cannot exceed the window; the other feature groups treat overlaps as defined in Appendix~\ref{app:features}. Giving overlapped time to the earlier event in the temporal gaze runs as well changes temporal and bigram features in 27 MapTask windows and none in MUNDEX, and no macro-F1 score by .001 or more. Coverage is the fraction of a window covered by valid canonical gaze; windows where either participant has coverage below 30\% are dropped (17 MapTask and 149 MUNDEX windows).
Tables~\ref{tab:app-maptask-windows} and \ref{tab:app-mundex-windows} vary the MapTask post-reference context (0--3\,s) and the MUNDEX symmetric context (1--3\,s) for the LR structured model; results are stable across these ranges.

\begin{table}[t]
\centering
\footnotesize
\setlength{\tabcolsep}{3pt}
\begin{tabular}{@{}lrrrr@{}}
\toprule
Window (s) & $n$ & Acc. & Macro-F1 & F1$_{-}$ \\
\midrule
0.000 & 5149 & 0.671 & 0.509 & 0.227 \\
1.000 & 5147 & 0.600 & 0.520 & 0.323 \\
1.500 & 5144 & 0.571 & 0.512 & 0.342 \\
2.000 & 5144 & 0.566 & 0.510 & 0.343 \\
3.000 & 5142 & 0.553 & 0.506 & 0.354 \\
\bottomrule
\end{tabular}

\caption{MapTask: prediction performance (macro-F1) of LR with the structured feature set under varying post-reference context lengths.}
\label{tab:app-maptask-windows}
\end{table}

\begin{table}[t]
\centering
\footnotesize
\setlength{\tabcolsep}{3pt}
\begin{tabular}{@{}lrrrr@{}}
\toprule
Window (s) & $n$ & Acc. & Macro-F1 & F1$_{-}$ \\
\midrule
1.000 & 804 & 0.544 & 0.543 & 0.536 \\
2.000 & 807 & 0.543 & 0.543 & 0.536 \\
3.000 & 807 & 0.549 & 0.549 & 0.552 \\
\bottomrule
\end{tabular}

\caption{MUNDEX: prediction performance (macro-F1) of LR with the structured feature set under varying symmetric context lengths.}
\label{tab:app-mundex-windows}
\end{table}

\section{Gaze Feature Definitions}
\label{app:features}

\providecommand{\featdef}[1]{\par\smallskip\noindent\textbf{#1}\enspace}

This appendix defines each gaze feature, grouped as in
Section~\ref{sec:experiments}. All features are computed from
discrete categories of gaze behaviors, not from eye-tracking fixations.

\subsection{Common definitions}
\label{app:features-common}

\featdef{Windows.} A window is $W=[a,b)$ with duration $T=b-a$ in
seconds. For MapTask, $a=s_{\mathrm{RE}}$ and $b=e_{\mathrm{RE}}+1.5$; for
MUNDEX, $a=\max(0,s_{\mathrm{UND}}-2)$ and $b=e_{\mathrm{UND}}+2$. Windows
are not truncated at the end of a recording, so unannotated time counts
towards $T$ but towards no gaze label.

\featdef{Gaze labels.} Gaze annotations are mapped to partner, task, or
away. MapTask \emph{up}, \emph{down}, and \emph{off} map to these categories
in both visibility conditions. In MUNDEX, TABLE maps to task, AWAY to
away, and EX or EE gaze targets to partner. Unassigned gaze remains
missing rather than being treated as away. Events with non-positive
duration are excluded.

\featdef{Participants.} Feature names carry a participant prefix:
\path{spk_}/\path{addr_} for the speaker and addressee of the MapTask
reference expression, and \path{ex_}/\path{ee_} for the MUNDEX explainer
and explainee, irrespective of who provided the understanding judgment.

\featdef{Observed durations.} Within each window, $D_k$ denotes the
observed duration of gaze category $k$, and $O=\sum_k D_k$ is the total
observed gaze time. Overlapping time is counted once and assigned to the
earlier-starting event. These durations underlie the participant gaze
proportions and structured features; gaze runs and joint sampling are
defined separately below.

\subsection{Raw proportions (7/8 features)}
\label{app:features-raw}

\featdef{Partner, task, and away proportion}
(\path{prop_partner}, \path{prop_task}, \path{prop_away}). The share of
the window spent on each label, $p_k=D_k/T$. Because proportions are
relative to the full window, they sum to coverage rather than 1.

\featdef{Mutual gaze} (\path{mutual_gaze} in MapTask,
\path{mutual_gaze_derived} in MUNDEX). With $U_{u,\text{partner}}$ the union of
participant $u$'s partner-directed intervals within $W$,
\[
 |U_{1,\text{partner}}\cap U_{2,\text{partner}}|/T.
\]
It depends on partner-directed annotations alone, regardless of any
overlapping task or away annotation.

\featdef{Explicit mutual gaze} (MUNDEX only,
\path{mutual_gaze_explicit}). The total duration of MUTUAL\_GAZE
annotations within $W$, divided by $T$. In the analysed MUNDEX windows,
it is nearly perfectly correlated with mutual gaze ($r$ near 1).

\smallskip
The six participant proportions and mutual gaze give 7 MapTask
features; explicit mutual gaze gives 8 for MUNDEX.

\subsection{Structured features (13/14 features)}

Structured features add three measures per participant to the raw
proportions.

\featdef{Coverage} (\path{coverage}). The share of the window with an
assigned gaze label, $C=O/T$. A window is retained only if both
participants have $C\geq .30$.

\featdef{Transition count} (\path{transitions}). The number of label
changes $\sum_{i=2}^{m}\mathbf 1[k_i\neq k_{i-1}]$ between the $m$
contributing intervals in temporal order (0 if $m<2$). Unannotated gaps
do not interrupt this sequence: different labels on either side of a gap
count as a transition.

\featdef{Entropy} (\path{entropy}). Duration-weighted Shannon entropy
over observed time,
\[
 H=-\sum_{k:D_k>0} q_k\log_2 q_k,\qquad q_k=D_k/O,
\]
in bits (maximum $\log_2 3$).

\subsection{Temporal dynamics (21 features per participant)}
\label{app:features-temporal}

\featdef{Gaze runs.} A gaze run is an interval assigned to a single gaze
category within the analysis window. Overlapping events with the same
label are combined; a later-starting event with a different label ends
the preceding interval, which is not resumed afterwards. Consecutive
same-label intervals separated by at most .01\,s are merged, including
the intervening gap; longer gaps remain unobserved. Let the resulting
runs be $(s_i,e_i,k_i)$, with durations $\ell_i=e_i-s_i$.

\featdef{Run count} (3 features). For each label $k$, the number of runs $n_k$.

\featdef{Mean and maximum run duration} (6 features). For each label,
$\sum_{i:k_i=k}\ell_i/n_k$ and $\max_{i:k_i=k}\ell_i$ in seconds, and 0 if
the label does not occur.

\featdef{Switch rate} (1 feature). The number of label changes between
successive runs, divided by $T$.

\featdef{Latency} (2 features). For partner and task, the relative onset of the
first run with that label, $(s_{\mathrm{first},k}-a)/T$, and 1 if the
label does not occur.

\featdef{Dominant, first, and last label} (9 features). One-hot indicators of the
label with the largest total run duration, of the first run label, and
of the last run label.

\subsection{Transition bigrams (6 features per participant)}

\featdef{Bigram proportion.} Repeated labels are collapsed in the
sequence of run labels (Appendix~\ref{app:features-temporal}), including
repeats separated by a gap. With $n_{jk}$ the number of changes from $j$
to $k$, each of the six ordered pairs $j\neq k$ receives
\[
 B_{jk}=n_{jk}\Big/\textstyle\sum_{j'\neq k'}n_{j'k'}.
\]
The proportions sum to 1 if at least one change occurs and are all 0
otherwise. Like the transition count and switch rate, bigrams count
changes across unannotated gaps without locating them within the gap.

\subsection{Coordination (6 joint features)}
\label{app:coordination}

\featdef{Joint sampling.} Gaze is sampled at approximately 10\,Hz, with
at least 20 equally spaced points per window. Only points at which both
participants have an assigned gaze label contribute to the joint
sequence $J$ of label pairs $(k_1,k_2)$. All six coordination features
are calculated over these jointly observed points rather than over the
full window duration.

\featdef{Mutual task gaze.} The proportion of $J$ with $k_1=k_2=\text{task}$. It
indicates that both participants look at the task space, not that they
inspect the same object or landmark.

\featdef{Mutual partner gaze.} The proportion of $J$ with $k_1=k_2=\text{partner}$.
Unlike mutual gaze (Appendix~\ref{app:features-raw}), it is relative to
jointly observed sample points rather than to the window.

\featdef{Gaze alignment.} The proportion of $J$ with $k_1=k_2$,
including joint away gaze.

\featdef{Complementary gaze.} The proportion of $J$ in which one
participant looks at the partner and the other at the task.

\featdef{Joint entropy.} $-\sum_{q_{jk}>0}q_{jk}\log_2 q_{jk}$ in bits,
where $q_{jk}$ is the relative frequency of the ordered pair $(j,k)$ in
$J$ (nine possible pairs).

\featdef{Partner coupling.} The Pearson correlation of the indicators
$\mathbf1[k_1=\text{partner}]$ and $\mathbf1[k_2=\text{partner}]$ over $J$, and 0 if either is
constant. It captures simultaneous coupling without a lag.

\subsection{Derived ratios (9 features)}

The first three ratios are computed for each participant from the raw
proportions; the asymmetries compare the two participants.

\featdef{Partner/task ratio.} The partner proportion relative to the
task proportion, with the denominator floored at .02 to avoid extreme
ratios when task gaze is near zero:
\[
 p_{\text{partner}}\big/\max(p_{\text{task}},\,.02).
\]

\featdef{Engagement.} $p_{\text{partner}}+p_{\text{task}}$.

\featdef{Task dominance.} $p_{\text{task}}-p_{\text{partner}}$.

\featdef{Partner, task, and entropy asymmetry.} The absolute differences
between the two participants' partner proportions, task proportions, and
entropies:
\begin{gather*}
 |p_{1,\text{partner}}-p_{2,\text{partner}}|,\qquad
 |p_{1,\text{task}}-p_{2,\text{task}}|,\\
 |H_1-H_2|.
\end{gather*}

\subsection{Feature sets, missing values, and overlaps}

\featdef{Feature sets.} Structured features include the raw proportions.
The extensions add 42 temporal, 12 bigram, 6 coordination, or 9 ratio
features to the structured set (55/56, 25/26, 19/20, or 22/23 features),
and all extended features combine these groups (82/83). Counts exclude
the control variables of the baseline models. Association analyses use
the structured features, and the reference-chain analyses a subset of
them.

\featdef{Missing values.} Missing gaze is neither interpolated nor
treated as a separate category. All extended features are defined in
every analysed window.

\featdef{Overlapping annotations.} Overlap handling differs across
duration-based measures, mutual gaze, gaze runs, and joint sampling, so
feature groups can assign overlapping time differently. A sensitivity
analysis applying the duration-based overlap rule to temporal and bigram
features changes macro-F1 by less than .001 in both corpora
(Appendix~\ref{app:windows}).

\section{Full Association Tests}
\label{app:associations}

Tables~\ref{tab:app-maptask-mw} and \ref{tab:app-mundex-mw} report Mann--Whitney U tests over all structured features; Table~\ref{tab:app-mundex-kruskal} reports Kruskal--Wallis tests across MUNDEX's four understanding labels. Positive class is aligned (MapTask) and annotator-judged UND (MUNDEX); $r$ is the rank-biserial correlation.

\begin{table}[t]
\centering
\footnotesize
\setlength{\tabcolsep}{3pt}
\begin{tabular}{@{}lrrrr@{}}
\toprule
Feature & Pos. & Neg. & $p$ & $r$ \\
\midrule
spk\_prop\_partner & 0.166 & 0.196 & $<.001$ & -0.057 \\
spk\_prop\_task & 0.805 & 0.773 & $<.001$ & 0.058 \\
spk\_entropy & 0.210 & 0.256 & $<.001$ & -0.054 \\
spk\_transitions & 0.434 & 0.526 & $<.001$ & -0.053 \\
addr\_entropy & 0.130 & 0.158 & 0.005 & -0.035 \\
addr\_transitions & 0.235 & 0.283 & 0.007 & -0.033 \\
addr\_prop\_away & 0.057 & 0.037 & 0.008 & 0.021 \\
addr\_prop\_partner & 0.081 & 0.094 & 0.014 & -0.030 \\
mutual\_gaze & 0.016 & 0.022 & 0.066 & -0.013 \\
spk\_coverage & 1.000 & 1.000 & 0.274 & 0.001 \\
addr\_prop\_task & 0.861 & 0.868 & 0.634 & 0.007 \\
addr\_coverage & 1.000 & 1.000 & 0.682 & 0.000 \\
spk\_prop\_away & 0.028 & 0.031 & 0.825 & -0.001 \\
\bottomrule
\end{tabular}

\caption{MapTask: Mann--Whitney U tests comparing aligned vs.\ non-aligned windows for each structured feature. Pos./Neg.: mean feature value in aligned/non-aligned windows; $r$: rank-biserial correlation (effect size; positive = higher in aligned windows); $p$: two-sided $p$-value. Table~\ref{tab:associations} selects the four features with the largest $|r|$ from each corpus.}
\label{tab:app-maptask-mw}
\end{table}

\begin{table}[t]
\centering
\footnotesize
\setlength{\tabcolsep}{3pt}
\begin{tabular}{@{}lrrrr@{}}
\toprule
Feature & Pos. & Neg. & $p$ & $r$ \\
\midrule
ex\_prop\_task & 0.683 & 0.579 & $<.001$ & 0.181 \\
ex\_prop\_partner & 0.258 & 0.360 & $<.001$ & -0.172 \\
mutual\_gaze\_explicit & 0.143 & 0.203 & $<.001$ & -0.143 \\
mutual\_gaze\_derived & 0.143 & 0.202 & $<.001$ & -0.143 \\
ee\_prop\_partner & 0.291 & 0.376 & $<.001$ & -0.148 \\
ee\_prop\_task & 0.620 & 0.535 & 0.001 & 0.131 \\
ex\_transitions & 1.297 & 1.620 & 0.004 & -0.113 \\
ex\_coverage & 0.962 & 0.954 & 0.005 & 0.115 \\
ee\_entropy & 0.376 & 0.461 & 0.005 & -0.110 \\
ee\_coverage & 0.943 & 0.937 & 0.006 & 0.112 \\
ee\_transitions & 1.217 & 1.432 & 0.011 & -0.098 \\
ex\_entropy & 0.459 & 0.504 & 0.121 & -0.062 \\
ex\_prop\_away & 0.020 & 0.015 & 0.510 & -0.014 \\
ee\_prop\_away & 0.031 & 0.027 & 0.935 & -0.002 \\
\bottomrule
\end{tabular}

\caption{MUNDEX: Mann--Whitney U tests comparing UND vs.\ non-UND windows. Columns as in Table~\ref{tab:app-maptask-mw}; Pos./Neg.\ are UND/non-UND means.}
\label{tab:app-mundex-mw}
\end{table}

\begin{table}[t]
\centering
\footnotesize
\setlength{\tabcolsep}{3pt}
\begin{tabular}{@{}lrr@{}}
\toprule
Feature & $H$ & $p$ \\
\midrule
ex\_prop\_task & 21.042 & $<.001$ \\
ex\_prop\_partner & 19.666 & $<.001$ \\
ex\_transitions & 15.512 & 0.001 \\
mutual\_gaze\_explicit & 14.926 & 0.002 \\
mutual\_gaze\_derived & 14.919 & 0.002 \\
ee\_prop\_partner & 14.758 & 0.002 \\
ee\_coverage & 13.320 & 0.004 \\
ee\_prop\_task & 12.840 & 0.005 \\
ee\_transitions & 11.692 & 0.009 \\
ee\_entropy & 10.254 & 0.017 \\
ex\_coverage & 9.253 & 0.026 \\
ex\_entropy & 7.750 & 0.051 \\
ex\_prop\_away & 4.799 & 0.187 \\
ee\_prop\_away & 2.525 & 0.471 \\
\bottomrule
\end{tabular}

\caption{MUNDEX: Kruskal--Wallis tests across all four understanding levels (UND, PART\_UND, NON\_UND, MISUND). Unlike the binary Mann--Whitney tests, this preserves the four-category distinction; a significant result indicates that at least one level differs, but does not test an ordinal trend.}
\label{tab:app-mundex-kruskal}
\end{table}

\section{Role- and Condition-Stratified Analyses}
\label{app:roles}

The pooled tests in Appendix~\ref{app:associations} combine all windows regardless of who produced or annotated them. Here we ask whether the gaze--grounding association differs by interactional role. Tables~\ref{tab:app-maptask-roles} and \ref{tab:app-mundex-roles} repeat the binary tests separately per role ($q$ is BH/FDR-adjusted within each stratum). In MapTask, giver-produced references show small but statistically significant effects ($|r|$ up to .086), while follower-produced references show near-zero effect sizes ($|r| \leq .043$). The follower stratum is smaller (1,698 vs.\ 3,446 giver windows). Subsampling giver expressions to the follower sample size 1{,}000 times, without preserving dialogue-level clustering, keeps rejection rates at .887--.930, while follower effects stay near zero ($|r| \leq .024$, $p \geq .31$). A feature$\times$role interaction in a logistic GEE for each structured feature does not survive BH correction in MapTask ($q{=}.064$) and is not significant in MUNDEX.

Table~\ref{tab:app-maptask-conditions} stratifies the MapTask tests by the corpus's visibility manipulation: in the eye-contact (ec) condition participants could see each other's faces, while in the no-eye-contact (nc) condition a barrier prevented it. Our gaze-annotated subset contains 31 ec dialogues (3,199 windows) and 15 nc dialogues (1,945 windows). The up $\rightarrow$ partner mapping is only literally partner-directed in the ec condition. All main associations are FDR-significant in the ec stratum with slightly larger effect sizes than in the pooled analysis, while in the nc stratum partner-directed gaze largely disappears (mean speaker partner gaze .04 vs.\ .25 in ec, averaged over all windows) and no feature reaches significance. This is consistent with the gaze--grounding signal coming from dialogues where partner-directed gaze is an available interactional resource, though a formal feature$\times$condition interaction does not reach significance.

\begin{table*}[t]
\centering
\footnotesize
\setlength{\tabcolsep}{3pt}
\begin{tabular}{@{}llrrrrr@{}}
\toprule
Role & Feature & Pos. & Neg. & $p$ & $q$ & $r$ \\
\midrule
follower & addr\_entropy & 0.169 & 0.202 & 0.063 & 0.412 & -0.043 \\
follower & addr\_transitions & 0.307 & 0.374 & 0.063 & 0.412 & -0.043 \\
follower & spk\_prop\_away & 0.066 & 0.062 & 0.293 & 0.788 & 0.016 \\
follower & spk\_prop\_partner & 0.127 & 0.129 & 0.311 & 0.788 & -0.024 \\
follower & addr\_prop\_partner & 0.147 & 0.143 & 0.457 & 0.788 & -0.019 \\
follower & addr\_prop\_away & 0.016 & 0.010 & 0.517 & 0.788 & 0.004 \\
follower & spk\_coverage & 1.000 & 1.000 & 0.536 & 0.788 & -0.001 \\
follower & addr\_coverage & 1.000 & 1.000 & 0.536 & 0.788 & -0.001 \\
follower & spk\_entropy & 0.198 & 0.220 & 0.582 & 0.788 & -0.014 \\
follower & addr\_prop\_task & 0.838 & 0.847 & 0.606 & 0.788 & 0.013 \\
follower & spk\_transitions & 0.390 & 0.420 & 0.673 & 0.796 & -0.010 \\
follower & spk\_prop\_task & 0.807 & 0.808 & 0.765 & 0.828 & 0.008 \\
follower & mutual\_gaze & 0.021 & 0.020 & 0.880 & 0.880 & 0.002 \\
\midrule
giver & spk\_prop\_task & 0.805 & 0.753 & $<.001$ & $<.001$ & 0.086 \\
giver & spk\_transitions & 0.455 & 0.583 & $<.001$ & $<.001$ & -0.076 \\
giver & spk\_entropy & 0.216 & 0.276 & $<.001$ & $<.001$ & -0.076 \\
giver & spk\_prop\_partner & 0.185 & 0.232 & $<.001$ & $<.001$ & -0.080 \\
giver & mutual\_gaze & 0.013 & 0.023 & 0.015 & 0.040 & -0.020 \\
giver & addr\_prop\_away & 0.077 & 0.052 & 0.019 & 0.041 & 0.026 \\
giver & addr\_prop\_partner & 0.050 & 0.068 & 0.040 & 0.074 & -0.027 \\
giver & addr\_entropy & 0.112 & 0.135 & 0.062 & 0.101 & -0.027 \\
giver & addr\_transitions & 0.200 & 0.234 & 0.082 & 0.111 & -0.025 \\
giver & spk\_prop\_away & 0.010 & 0.014 & 0.086 & 0.111 & -0.008 \\
giver & spk\_coverage & 1.000 & 0.999 & 0.097 & 0.115 & 0.002 \\
giver & addr\_coverage & 1.000 & 0.999 & 0.444 & 0.481 & 0.001 \\
giver & addr\_prop\_task & 0.873 & 0.880 & 0.945 & 0.945 & -0.001 \\
\bottomrule
\end{tabular}

\caption{MapTask: Mann--Whitney tests stratified by the role that produced the reference expression. Giver-produced references ($n{=}3{,}446$) show small but statistically significant effects ($|r|$ up to .086, $q{<}.001$); follower-produced references ($n{=}1{,}698$) show near-zero effect sizes ($|r| \leq .043$, no $q{<}.05$). $q$: BH-corrected within each stratum.}
\label{tab:app-maptask-roles}
\end{table*}

\begin{table*}[t]
\centering
\footnotesize
\setlength{\tabcolsep}{3pt}
\begin{tabular}{@{}llrrrrr@{}}
\toprule
Role & Feature & Pos. & Neg. & $p$ & $q$ & $r$ \\
\midrule
EE & ex\_prop\_partner & 0.270 & 0.358 & 0.016 & 0.068 & -0.151 \\
EE & ex\_prop\_task & 0.669 & 0.580 & 0.019 & 0.068 & 0.149 \\
EE & mutual\_gaze\_derived & 0.141 & 0.204 & 0.019 & 0.068 & -0.140 \\
EE & mutual\_gaze\_explicit & 0.141 & 0.204 & 0.019 & 0.068 & -0.140 \\
EE & ee\_coverage & 0.942 & 0.935 & 0.031 & 0.085 & 0.137 \\
EE & ee\_prop\_partner & 0.287 & 0.371 & 0.036 & 0.085 & -0.130 \\
EE & ee\_prop\_task & 0.621 & 0.538 & 0.046 & 0.091 & 0.127 \\
EE & ex\_transitions & 1.393 & 1.743 & 0.056 & 0.098 & -0.117 \\
EE & ex\_coverage & 0.956 & 0.954 & 0.257 & 0.400 & 0.072 \\
EE & ee\_entropy & 0.398 & 0.442 & 0.410 & 0.574 & -0.050 \\
EE & ex\_prop\_away & 0.017 & 0.015 & 0.462 & 0.588 & -0.024 \\
EE & ee\_transitions & 1.200 & 1.290 & 0.571 & 0.666 & -0.034 \\
EE & ex\_entropy & 0.501 & 0.515 & 0.676 & 0.729 & -0.026 \\
EE & ee\_prop\_away & 0.034 & 0.026 & 0.901 & 0.901 & -0.005 \\
\midrule
EX & ex\_prop\_task & 0.692 & 0.577 & $<.001$ & 0.002 & 0.206 \\
EX & ex\_prop\_partner & 0.251 & 0.362 & $<.001$ & 0.003 & -0.186 \\
EX & ee\_prop\_partner & 0.294 & 0.380 & 0.002 & 0.007 & -0.165 \\
EX & ee\_transitions & 1.227 & 1.562 & 0.002 & 0.007 & -0.156 \\
EX & ee\_entropy & 0.362 & 0.478 & 0.002 & 0.007 & -0.157 \\
EX & mutual\_gaze\_explicit & 0.144 & 0.203 & 0.004 & 0.009 & -0.145 \\
EX & mutual\_gaze\_derived & 0.144 & 0.201 & 0.004 & 0.009 & -0.144 \\
EX & ee\_prop\_task & 0.620 & 0.533 & 0.012 & 0.021 & 0.135 \\
EX & ex\_coverage & 0.965 & 0.954 & 0.016 & 0.024 & 0.130 \\
EX & ex\_transitions & 1.240 & 1.506 & 0.055 & 0.077 & -0.099 \\
EX & ee\_coverage & 0.944 & 0.940 & 0.064 & 0.082 & 0.100 \\
EX & ex\_entropy & 0.434 & 0.494 & 0.131 & 0.153 & -0.079 \\
EX & ex\_prop\_away & 0.022 & 0.015 & 0.825 & 0.889 & -0.006 \\
EX & ee\_prop\_away & 0.030 & 0.027 & 0.922 & 0.922 & -0.003 \\
\bottomrule
\end{tabular}

\caption{MUNDEX: Mann--Whitney tests stratified by annotator role. EE: the explainee's own understanding report; EX: the explainer's judgment of the explainee's understanding. $q$: BH-corrected within each stratum.}
\label{tab:app-mundex-roles}
\end{table*}

\begin{table*}[t]
\centering
\footnotesize
\setlength{\tabcolsep}{3pt}
\begin{tabular}{@{}llrrrrr@{}}
\toprule
Cond. & Feature & Pos. & Neg. & $p$ & $q$ & $r$ \\
\midrule
ec & spk\_entropy & 0.316 & 0.379 & $<.001$ & 0.002 & -0.078 \\
ec & spk\_prop\_task & 0.728 & 0.685 & $<.001$ & 0.002 & 0.078 \\
ec & spk\_transitions & 0.652 & 0.773 & $<.001$ & 0.002 & -0.073 \\
ec & addr\_prop\_away & 0.049 & 0.021 & $<.001$ & 0.003 & 0.029 \\
ec & spk\_prop\_partner & 0.243 & 0.280 & 0.001 & 0.003 & -0.070 \\
ec & addr\_prop\_partner & 0.101 & 0.126 & 0.004 & 0.008 & -0.051 \\
ec & addr\_entropy & 0.180 & 0.215 & 0.012 & 0.023 & -0.044 \\
ec & addr\_transitions & 0.330 & 0.391 & 0.020 & 0.033 & -0.041 \\
ec & mutual\_gaze & 0.025 & 0.035 & 0.072 & 0.104 & -0.020 \\
ec & spk\_coverage & 1.000 & 0.999 & 0.112 & 0.146 & 0.002 \\
ec & addr\_coverage & 1.000 & 0.999 & 0.282 & 0.324 & 0.002 \\
ec & spk\_prop\_away & 0.028 & 0.035 & 0.299 & 0.324 & -0.008 \\
ec & addr\_prop\_task & 0.850 & 0.852 & 0.360 & 0.360 & 0.017 \\
\midrule
nc & spk\_prop\_partner & 0.041 & 0.054 & 0.066 & 0.737 & -0.025 \\
nc & addr\_transitions & 0.079 & 0.101 & 0.201 & 0.737 & -0.017 \\
nc & addr\_entropy & 0.049 & 0.062 & 0.203 & 0.737 & -0.017 \\
nc & spk\_prop\_away & 0.028 & 0.025 & 0.336 & 0.737 & 0.010 \\
nc & spk\_prop\_task & 0.931 & 0.921 & 0.343 & 0.737 & 0.016 \\
nc & addr\_coverage & 1.000 & 1.000 & 0.408 & 0.737 & -0.001 \\
nc & addr\_prop\_task & 0.880 & 0.895 & 0.464 & 0.737 & -0.014 \\
nc & spk\_transitions & 0.080 & 0.107 & 0.548 & 0.737 & -0.007 \\
nc & spk\_coverage & 1.000 & 1.000 & 0.559 & 0.737 & -0.001 \\
nc & spk\_entropy & 0.039 & 0.048 & 0.567 & 0.737 & -0.007 \\
nc & addr\_prop\_partner & 0.049 & 0.041 & 0.663 & 0.751 & 0.005 \\
nc & addr\_prop\_away & 0.071 & 0.064 & 0.694 & 0.751 & 0.006 \\
nc & mutual\_gaze & 0.001 & 0.000 & 0.981 & 0.981 & 0.000 \\
\bottomrule
\end{tabular}

\caption{MapTask: association tests stratified by the eye-contact (ec) versus no-eye-contact (nc) condition; $q$ is BH/FDR-adjusted within each condition.}
\label{tab:app-maptask-conditions}
\end{table*}

\section{Cluster-Robust Marginal Tests}
\label{app:gee}

Mann--Whitney tests treat windows as independent, although windows within a dialogue share participants and topic. To account for within-cluster dependence, Tables~\ref{tab:app-maptask-gee} and \ref{tab:app-mundex-gee} report one logistic GEE \citep{liang1986longitudinal} per standardized feature, using exchangeable within-cluster correlation and dialogue/interaction clusters. Features significant after BH correction are marked with $^\dagger$ in Table~\ref{tab:associations}. We use univariate models because the structured features are strongly collinear: label proportions sum to coverage, entropy tracks transitions, and explicit and derived mutual gaze are nearly identical. Joint coefficients would therefore be difficult to interpret.

Participants also recur beyond dialogues and interactions. The 46 MapTask dialogues involve 24 participants, who form six groups of dialogues connected by shared participants; eight of the nine MUNDEX explainers take part in three interactions and one in two. We refit each GEE with these groups or with explainers as clusters. Because robust standard errors are anti-conservative with few clusters, we also apply bias-reduced (Mancl--DeRouen) standard errors \citep{mancl2001covariance}. With dialogue clusters, the correction leaves MapTask speaker entropy and transitions significant ($q{=}.009$ and $q{=}.010$); with participant groups, no MapTask feature survives BH correction (smallest $q{=}.11$; speaker entropy $q{=}.23$). With explainer clusters and the correction, explainer task and partner gaze ($q{=}.001$ and $q{=}.028$) and explainee entropy and transitions (both $q{=}.039$) remain significant, whereas explainee task and partner gaze and both mutual-gaze measures do not ($q{=}.055$--$.062$). Coefficient signs do not change. The MapTask associations and the MUNDEX associations for explainee gaze proportions and mutual gaze are therefore not robust to treating recurring participants as the unit of inference.

\begin{table*}[t]
\centering
\footnotesize
\setlength{\tabcolsep}{3pt}
\begin{tabular}{@{}lrrrr@{}}
\toprule
Feature & Coef. & SE & $p$ & $q$ \\
\midrule
spk\_entropy & -0.108 & 0.031 & $<.001$ & 0.006 \\
spk\_transitions & -0.094 & 0.029 & 0.001 & 0.007 \\
addr\_prop\_away & 0.104 & 0.040 & 0.008 & 0.036 \\
spk\_prop\_task & 0.088 & 0.042 & 0.035 & 0.091 \\
addr\_entropy & -0.074 & 0.036 & 0.037 & 0.091 \\
spk\_prop\_partner & -0.084 & 0.041 & 0.042 & 0.091 \\
addr\_transitions & -0.066 & 0.035 & 0.061 & 0.112 \\
mutual\_gaze & -0.053 & 0.031 & 0.088 & 0.143 \\
spk\_coverage & 0.033 & 0.022 & 0.141 & 0.204 \\
addr\_prop\_partner & -0.060 & 0.048 & 0.215 & 0.279 \\
spk\_prop\_away & -0.018 & 0.029 & 0.539 & 0.637 \\
addr\_prop\_task & -0.018 & 0.043 & 0.679 & 0.687 \\
addr\_coverage & 0.011 & 0.028 & 0.687 & 0.687 \\
\bottomrule
\end{tabular}

\caption{MapTask: per-feature logistic GEE with dialogue-level clustering. Each row is a separate univariate model. Coef.: standardized log-odds (positive = higher odds of aligned per SD increase); SE: cluster-robust standard error; $q$: BH-corrected. Features with $q{<}.05$ are marked $^\dagger$ in Table~\ref{tab:associations}.}
\label{tab:app-maptask-gee}
\end{table*}

\begin{table*}[t]
\centering
\footnotesize
\setlength{\tabcolsep}{3pt}
\begin{tabular}{@{}lrrrr@{}}
\toprule
Feature & Coef. & SE & $p$ & $q$ \\
\midrule
ex\_prop\_partner & -0.351 & 0.075 & $<.001$ & $<.001$ \\
ex\_prop\_task & 0.319 & 0.068 & $<.001$ & $<.001$ \\
mutual\_gaze\_explicit & -0.266 & 0.077 & $<.001$ & 0.002 \\
mutual\_gaze\_derived & -0.265 & 0.078 & $<.001$ & 0.002 \\
ee\_prop\_partner & -0.240 & 0.082 & 0.004 & 0.009 \\
ee\_prop\_task & 0.234 & 0.081 & 0.004 & 0.009 \\
ee\_entropy & -0.195 & 0.070 & 0.005 & 0.011 \\
ee\_transitions & -0.180 & 0.069 & 0.009 & 0.016 \\
ex\_transitions & -0.206 & 0.082 & 0.012 & 0.019 \\
ex\_entropy & -0.128 & 0.077 & 0.095 & 0.132 \\
ex\_coverage & 0.094 & 0.075 & 0.209 & 0.266 \\
ee\_coverage & 0.074 & 0.072 & 0.310 & 0.361 \\
ex\_prop\_away & 0.054 & 0.092 & 0.556 & 0.599 \\
ee\_prop\_away & -0.023 & 0.063 & 0.709 & 0.709 \\
\bottomrule
\end{tabular}

\caption{MUNDEX: per-feature logistic GEE with interaction-level clustering. Columns as in Table~\ref{tab:app-maptask-gee}; positive coefficient = higher odds of UND per SD increase.}
\label{tab:app-mundex-gee}
\end{table*}

\section{Reference-Chain Analyses}
\label{app:chains}

To examine gaze change within a dialogue, we group mentions of the same landmark concept into chains, ordered by onset: 636 chains, of which 587 have more than one mention.

Table~\ref{tab:app-chain-positions} summarizes alignment and speaker gaze by mention position. For the resolution analysis (Table~\ref{tab:app-resolution}), we locate in each chain the first non-aligned mention followed by a later aligned mention of the same concept, and compare gaze at the last non-aligned mention with gaze at the resolving aligned mention (paired Wilcoxon, BH-corrected). The primary analysis restricts to same-speaker pairs, where the same person produced both mentions ($n{=}189$); only entropy survives correction ($q{=}.044$). Its standardized change is $d_z{=}{-}.20$, with a dialogue-cluster bootstrap 95\% CI of $[-.34, -.07]$. As a more conservative sensitivity check, averaging pair differences within each dialogue ($n{=}45$) yields $q{=}.20$ for entropy. Resampling the six participant groups of Appendix~\ref{app:gee} instead of dialogues gives a 95\% CI of $[-.37, -.01]$, but the mean entropy change is $-.28$ and $-.15$ in two groups and between $-.04$ and $.01$ in the other four, and a Wilcoxon test over the six group means gives $p{=}.16$. The result is therefore sensitive to the inference unit. The mixed-speaker pool ($n{=}328$, which includes cross-speaker resolutions) is reported as an additional sensitivity check. Table~\ref{tab:app-position-matched} repeats the aligned/non-aligned contrast within each mention-position bucket.

\begin{table*}[t]
\centering
\footnotesize
\setlength{\tabcolsep}{3pt}
\begin{tabular}{@{}lrrrrrrr@{}}
\toprule
Mention & $n$ & Aligned rate & Spk.\ partner & Spk.\ task & Spk.\ entropy & Spk.\ trans. & Mutual \\
\midrule
1 & 636 & 0.303 & 0.230 & 0.744 & 0.310 & 0.643 & 0.019 \\
2 & 587 & 0.595 & 0.172 & 0.797 & 0.215 & 0.463 & 0.022 \\
3 & 518 & 0.736 & 0.188 & 0.791 & 0.210 & 0.434 & 0.018 \\
4+ & 3403 & 0.847 & 0.162 & 0.808 & 0.209 & 0.426 & 0.016 \\
\bottomrule
\end{tabular}

\caption{MapTask: how alignment and speaker gaze evolve across repeated mentions of the same landmark. Each row is a mention position (1st, 2nd, \ldots); aligned rate is the proportion of mentions at that position that are aligned. Later positions generally show higher alignment and lower partner gaze, though the trend is not strictly monotonic.}
\label{tab:app-chain-positions}
\end{table*}

\begin{table*}[t]
\centering
\footnotesize
\setlength{\tabcolsep}{3pt}
\begin{tabular}{@{}lrrrrr@{}}
\toprule
Feature & Pre & Post & $\Delta$ & $p$ & $q$ \\
\midrule
spk\_entropy & 0.355 & 0.263 & -0.092 & 0.005 & 0.044 \\
spk\_prop\_partner & 0.239 & 0.187 & -0.052 & 0.020 & 0.079 \\
spk\_prop\_task & 0.755 & 0.800 & 0.045 & 0.052 & 0.138 \\
spk\_transitions & 0.720 & 0.608 & -0.111 & 0.208 & 0.407 \\
addr\_prop\_task & 0.852 & 0.874 & 0.023 & 0.255 & 0.407 \\
mutual\_gaze & 0.030 & 0.023 & -0.007 & 0.445 & 0.529 \\
addr\_entropy & 0.157 & 0.139 & -0.018 & 0.463 & 0.529 \\
addr\_prop\_partner & 0.093 & 0.081 & -0.012 & 0.608 & 0.608 \\
\bottomrule
\end{tabular}

\caption{MapTask: gaze at the last non-aligned mention (Pre) vs.\ the first subsequent aligned mention (Post) of the same concept; same-speaker pairs only (paired Wilcoxon, $n{=}189$; $q$ is BH-corrected). A separate mixed-speaker analysis ($n{=}328$) was examined as a sensitivity check: the four speaker features change in the same directions and all survive correction, whereas three addressee features change in the opposite direction; no addressee feature survives correction in either analysis.}
\label{tab:app-resolution}
\end{table*}

\begin{table*}[t]
\centering
\footnotesize
\setlength{\tabcolsep}{3pt}
\begin{tabular}{@{}llrrrrr@{}}
\toprule
Mention & Feature & Aligned mean & Non-al.\ mean & $p$ & $q$ & $r$ \\
\midrule
1 & spk\_transitions & 0.554 & 0.682 & 0.114 & 0.200 & -0.069 \\
1 & spk\_prop\_partner & 0.206 & 0.240 & 0.115 & 0.200 & -0.070 \\
1 & spk\_entropy & 0.269 & 0.327 & 0.150 & 0.200 & -0.063 \\
1 & spk\_prop\_task & 0.763 & 0.736 & 0.260 & 0.260 & 0.051 \\
\midrule
2 & spk\_prop\_task & 0.836 & 0.741 & 0.002 & 0.010 & 0.126 \\
2 & spk\_prop\_partner & 0.150 & 0.204 & 0.059 & 0.118 & -0.076 \\
2 & spk\_entropy & 0.200 & 0.238 & 0.166 & 0.222 & -0.054 \\
2 & spk\_transitions & 0.458 & 0.471 & 0.352 & 0.352 & -0.036 \\
\midrule
3 & spk\_prop\_task & 0.772 & 0.845 & 0.083 & 0.206 & -0.085 \\
3 & spk\_prop\_partner & 0.205 & 0.140 & 0.103 & 0.206 & 0.079 \\
3 & spk\_transitions & 0.449 & 0.394 & 0.478 & 0.638 & 0.033 \\
3 & spk\_entropy & 0.207 & 0.218 & 0.731 & 0.731 & 0.016 \\
\midrule
4+ & spk\_prop\_partner & 0.161 & 0.169 & 0.351 & 0.708 & -0.021 \\
4+ & spk\_prop\_task & 0.809 & 0.800 & 0.443 & 0.708 & 0.018 \\
4+ & spk\_transitions & 0.422 & 0.453 & 0.605 & 0.708 & -0.011 \\
4+ & spk\_entropy & 0.208 & 0.214 & 0.708 & 0.708 & -0.008 \\
\bottomrule
\end{tabular}

\caption{MapTask: aligned vs.\ non-aligned within each mention-position bucket (Mann--Whitney). Stratifying by position addresses the confound that later mentions are both more aligned and more task-directed. $q$ is BH-corrected within each position bucket (4 tests per bucket). Only task-gaze proportion at second mentions survives correction ($q{=}.01$); first-mention contrasts are not significant (aligned rate .303).}
\label{tab:app-position-matched}
\end{table*}

\section{Prediction Setup and Full Results}
\label{app:prediction}

As a complementary signal check, we test whether the gaze features defined in Appendix~\ref{app:features} carry enough information to predict grounding state. All classifiers are scikit-learn~\citep{scikit-learn} logistic regressions (standardized features, balanced class weights, \texttt{max\_iter}=1000). Cross-validation is GroupKFold with 10 splits grouped by dialogue (MapTask) and 5 splits grouped by explainer (MUNDEX, held-out-explainer), so that MUNDEX test-fold performance reflects generalization to an unseen explainer. Controls are condition and speaker-role indicators (MapTask) and annotator-role (MUNDEX). Tables~\ref{tab:app-maptask-models} and \ref{tab:app-mundex-models} give the full ablation with per-class F1.

These scores come from one deterministic GroupKFold partition. Repeating the controls-only model and every feature group over 30 reshuffled grouped partitions with the same numbers of folds shows how much the scores depend on the partition. In MapTask, the controls-only score ranges from .460 to .511 (mean .492); structured+temporal features score highest in 27 of the 30 partitions (all extended in the other three), and their gain over controls ranges from .015 to .070 (mean .038). In MUNDEX, the role-only control does not vary across partitions; raw proportions score highest in 27 partitions, and their gain over the control is at most .027 (mean .017) and not positive in one partition.

\begin{table*}[t]
\centering
\footnotesize
\setlength{\tabcolsep}{3pt}
\begin{tabular}{@{}lrrrrr@{}}
\toprule
Model / feature set & \#feat. & Acc. & Macro-F1 & F1$_{+}$ & F1$_{-}$ \\
\midrule
Majority baseline & -- & 0.740 & 0.425 & 0.851 & 0.000 \\
LR (controls only) & -- & 0.514 & 0.472 & 0.622 & 0.322 \\
LR raw proportions & 7 & 0.594 & 0.522 & 0.708 & 0.336 \\
LR structured & 13 & 0.571 & 0.512 & 0.682 & 0.342 \\
~~~+ temporal & 55 & 0.621 & 0.532 & 0.736 & 0.328 \\
~~~+ bigrams & 25 & 0.596 & 0.521 & 0.710 & 0.332 \\
~~~+ coordination & 19 & 0.573 & 0.511 & 0.685 & 0.338 \\
~~~+ ratios & 22 & 0.584 & 0.516 & 0.697 & 0.335 \\
LR all extended & 82 & 0.618 & 0.527 & 0.734 & 0.319 \\
\bottomrule
\end{tabular}

\caption{MapTask: full prediction results (aligned vs.\ non-aligned). F1$_{+}$/F1$_{-}$ are the aligned/non-aligned classes; ``$+$'' rows add one feature group to the LR structured set.}
\label{tab:app-maptask-models}
\end{table*}

\begin{table*}[t]
\centering
\footnotesize
\setlength{\tabcolsep}{3pt}
\begin{tabular}{@{}lrrrrr@{}}
\toprule
Model / feature set & \#feat. & Acc. & Macro-F1 & F1$_{+}$ & F1$_{-}$ \\
\midrule
Majority baseline & -- & 0.554 & 0.356 & 0.000 & 0.713 \\
LR (controls only) & -- & 0.544 & 0.544 & 0.550 & 0.538 \\
LR raw proportions & 8 & 0.564 & 0.564 & 0.560 & 0.568 \\
LR structured & 14 & 0.543 & 0.543 & 0.549 & 0.536 \\
~~~+ temporal & 56 & 0.551 & 0.550 & 0.529 & 0.572 \\
~~~+ bigrams & 26 & 0.523 & 0.522 & 0.495 & 0.548 \\
~~~+ coordination & 20 & 0.551 & 0.551 & 0.556 & 0.546 \\
~~~+ ratios & 23 & 0.545 & 0.545 & 0.563 & 0.526 \\
LR all extended & 83 & 0.527 & 0.527 & 0.519 & 0.534 \\
\bottomrule
\end{tabular}

\caption{MUNDEX: full prediction results (UND vs.\ non-UND).}
\label{tab:app-mundex-models}
\end{table*}

\end{document}